\documentclass[10pt,twocolumn,letterpaper]{article}

\usepackage{times}
\usepackage{epsfig}
\usepackage{graphicx}
\usepackage{amsmath}
\usepackage{amssymb}
\usepackage{algorithm2e}
\usepackage{algpseudocode}
\usepackage{array}

\begin{document}

\title{Learning Rotation for Kernel Correlation Filter}

\author{Abdullah Hamdi, Bernard Ghanem \\
King Abdullah University of Science and Technology (KAUST), Thuwal, Saudi Arabia \\
{\tt\small \{abdullah.hamdi, Bernard.Ghanem\} @kaust.edu.sa}
}
\maketitle

\begin{abstract}
Kernel Correlation Filters have shown a very promising scheme for visual tracking in terms of speed and accuracy on several benchmarks. However it suffers from problems that affect its performance like occlusion, rotation and scale change. This paper tries to tackle the problem of rotation by reformulating the optimization problem for learning the correlation filter. This modification (RKCF) includes learning rotation filter that utilizes circulant structure of HOG feature to guesstimate rotation from one frame to another and enhance the detection of KCF. Hence it gains boost in overall accuracy in many of OBT50 detest videos with minimal additional computation
\end{abstract}

\section{Introduction}
Visual object tracking is a very important task in computer vision in which an object of interest would be identified and located in the first frame of a video. The goal is to follow the object movement and scale in subsequent frames by applying the tracking algorithm, usually faster and more efficient then a general detection scheme. Because of the wide range of applications that include visual tracking (e.g. robotics and surveillance), tracking had the attention of the computer vision community for several years\\
Correlation filters (CF) have used in tracking for several years in visual tracking due to their speed and efficient computations .Adding Kernels to these trackers produced state of the art Kernel Correlation Filter (KCF) that topped tracking benchmarks for several years. KCF utilizes the circulant structure of the data matrix of all possible shifts to achieve less computation and utilizes the Kernel substitution trick  \cite{Author1}\\
Several versions and modifications of KCF came to tackle the problems it possessed like occlusion,boundary effect,scale,and rotation \cite{Author3} \cite{Author4}\cite{Author5}\cite{Author6}.\\
Rotation being one of these problems that KCF suffers is itself an interesting problem (rotation detection) with wide range of applications like texture classification \cite{Author7}. By the way KCF filter is constructed it assumes the object didn't rotate or change shape, this assumption cause the response of the filter to deteriorate and the detection would (as a result) drift away from the target position and cause drop of the performance.\\
We propose here to reformulate the optimization objective to include a second filter that will learn a rotation descriptor for the target and utilize this information in the detection phase of KCF family tracker. We show that this extra information ( the rotation from one frame to another ) will enhance the performance of CF trackers with many of its variations , with potential of applying in trackers that utilize deep features and Siamese network as discussed in \cite{Author5}. This rotation filter uses the circularity of the HOG feature that enable it to utilize the same computational efficiency of the KCF tracker (as we will show in section 3) giving boost to the base performance with almost no additional computation time.

\section{Related work}
\subsection{Kernel Correlation Filter}
\begin{figure}[h!]
\begin{center}
   \includegraphics[width=0.8\linewidth]{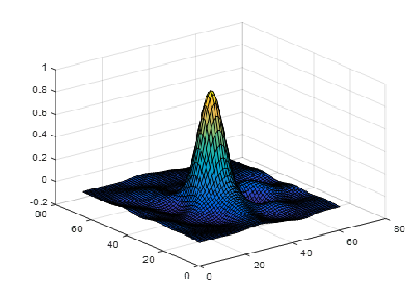}
\end{center}
   \caption{typical response of applying the learned KCF filter on an image patch}
\label{surf}
\end{figure}

State of the art tracking technique that utilizes the cyclic nature of the shifted patches and Kernel trick to implement a very fast tracking algorithm based on correlation filters. They are filters that try to guesstimate the new position of the target by learning  filters on each patch with expected response to be Gaussian with maximum at the center if the object didn't move .Figure \ref{surf} shows a typical response/target of regression of the filter \\
\\

If the object moved by little from frame to another , the maximum response will be shifted and the translation of the maximum will be used to translate the patch in the image , and so forth in the following frames. \cite{Author1}. To achieve this ,a filter $ \mathbf{w} $ that minimize the energy of error between the response of the filter on patch image and a typical response of stationary patch according to the following equation

\begin{equation} \label{eq:1}
\min_{\mathbf{w}} \parallel \mathbf{X}\mathbf{w} - \mathbf{y} \parallel_{2}^{2} + \lambda \parallel \mathbf{w} \parallel_{2}^{2} 
\end{equation}

where $\mathbf{y}$ is Gaussian  response of the filter if the patch didn't move and $ \mathbf{X} $ is the data matrix of the image patch that contains the target being tracked , $ \lambda $ is the regularization variable. \\
the closed form solution of the optimization \ref{eq:1} given by :

\begin{equation}\label{eq:2}
 \boldsymbol{w} = (\boldsymbol{X}^{T} \boldsymbol{X} + \lambda\boldsymbol{I} ) ^{-1} \boldsymbol{X}^{T}\boldsymbol{y} 
\end{equation}

It can be seen that the matrix $\mathbf{X}$ is a circulant matrix of all possible shifts of the vector $\mathbf{x}$ which is the vectorized image patch surrounding the target being tracked. Since the matrix $\mathbf{X}$ is circulant it can diagonalized  by the DFT matrix  as follows :

\begin{equation}\label{eq:3}
\boldsymbol{X} = \boldsymbol{F}diag(\boldsymbol{\hat{x}})\boldsymbol{F}^{H}
\end{equation}

where $\mathbf{\hat{x}} $ is DFT of vector $ \mathbf{x} $  .  Using \ref{eq:3} we can see that 
\begin{equation}\label{eq:4}
\boldsymbol{X}^{H}\boldsymbol{X} = \boldsymbol{F}diag(\boldsymbol{\hat{x}^{*}} \odot \boldsymbol{\hat{x}})\boldsymbol{F}^{H}
\end{equation}
substituting \ref{eq:4} into \ref{eq:2} we get the following closed form solution to the learned filter $\mathbf{\hat{w}} $ in the Fourier domain in each frame.
 
 \begin{equation}\label{eq:5}
 \hat{\boldsymbol{w}} = \frac{\hat{\boldsymbol{x}} \bigodot \hat{\boldsymbol{y}}}{\hat{\boldsymbol{x}} \bigodot \hat{\boldsymbol{x}} + \boldsymbol{\lambda}}
\end{equation}  

The hat indicate DFT of the term. The solution in dual domain is : \\ 
 \begin{equation}\label{eq:6}
\hat{\boldsymbol{\alpha }} = \frac{ \hat{\boldsymbol{y}}}{\hat{\boldsymbol{x}} \bigodot \hat{\boldsymbol{y}} + \boldsymbol{\lambda}}
 \end{equation} 
 
if we added the Kernels and used the Kernel trick we can show that the dual will have the form 
\begin{equation}\label{eq:7}
    \boldsymbol{\hat{\alpha}} = \frac{\boldsymbol{\hat{y}}}{\boldsymbol{\hat{k}^{xx}}  + \lambda} 
\end{equation}
where $\mathbf{\hat{k}^{xx}}$ is the kernel vector formed from inner product from $\mathbf{x} $ and itself

The KCF take advantage of the fact that the convolution of two patches (loosely, their dot-product at different relative translations) is equivalent to an element-wise product in the Fourier domain. Thus, by formulating their objective in the Fourier domain, they can specify the desired output of a linear classifier for several translations, or image shifts, at once\cite{Author1}. The power of the kernel trick comes from the implicit use of a high-dimensional feature space , without ever instantiating a vector in that space.\cite{Author1}

\subsection{KCF Family of trackers }
One adaption of KCF is SAMF (Scale-Adaptive Kernel Correlation Filter) \cite{Author3}. This adaption just like KCF , learn the filter and apply it on translated patches , however it searched for different scales and look at the maximum response over all the scales . We are solving the rotation problem it more efficient way (one shot) rather than trying all different rotations and take the one with the highest response.\\
other versions of KCF are those of deep features that enhance the performance of KCF and allow it to be scale and rotational invariant for enough training of deep Neural Network like \cite{Author5}. However these requires long training and GPUs and also lack the speed the original KCF has.

\subsection{Histogram of oriented gradients ( HOG }
\begin{figure}[h!]
\begin{center}
   \includegraphics[width=0.8\linewidth]{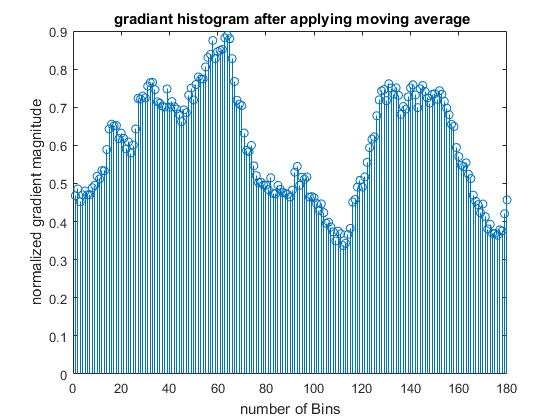}
\end{center}
   \caption{HOG feature of image patch , smoothed for better performance ,  observe the circular structure of the global descriptor }
\label{hog}
\end{figure}

\begin{figure}[h!]
\begin{center}
   \includegraphics[width=0.8\linewidth]{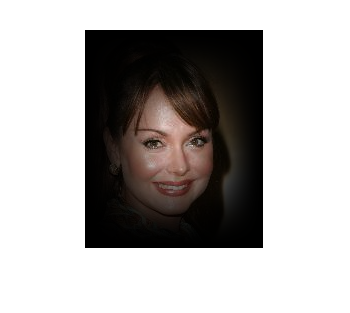}
\end{center}
   \caption{image patch multiplied by cos window, (typical in CF trackers)  to reduce the effect of background and the boundary effect }
\label{patch}
\end{figure}

\begin{figure}[t]
\begin{center}
   \includegraphics[width=0.8\linewidth]{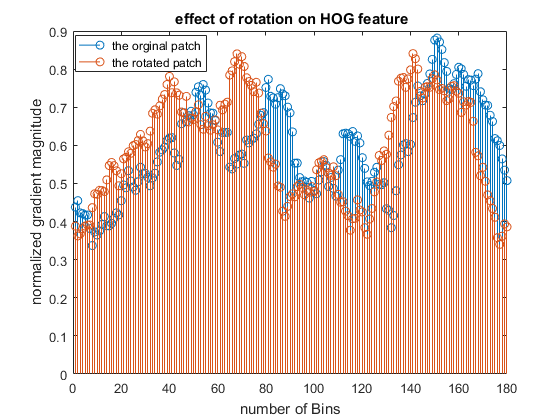}
\end{center}
   \caption{the effect of rotating image like the one in \ref{patch} on its global HOG descriptor }
\label{effect}
\end{figure}

A generic way to extract orientation feature of objects is to find the distribution of the gradients in cells that combine a number of pixels. This is very powerful and fast technique to characterize the orientation of an object. If we take the image patch to be one cell and we choose enough number of bins we can have a global descriptor for the patch as can be seen in figure \ref{hog}, it can be seen that the descriptor has circular structure ( coming from the fact that rotating 180 degrees give the same HOG descriptor for the patch) that will prove crucial in formulating the solution for RKCF.

 for an image patch like the one in \ref{patch} by multiplying it by a cos window and then rotating it, its global HOG descriptor will suffer a shift like the one in figure \ref{effect} which is very similar to what will happen to the target when we learn the rotation filter and apply it on the new rotated patch in RKCF

\space 
\section{methodology}

\subsection{Derivation of augmented KCF ( RKCF)}
We can extend the optimizing objective of KCF ( or any other tracker that uses Correlation Filters ) to include the rotation information of the target as follows
\space \\  
\begin{equation} \label{eq:8}
\min_{\boldsymbol{w},\boldsymbol{r}} \parallel \boldsymbol{X}\boldsymbol{w} - \boldsymbol{y} \parallel_{2}^{2} + \lambda_{1} \parallel \boldsymbol{w} \parallel_{2}^{2} + \parallel \boldsymbol{A}\boldsymbol{r} - \boldsymbol{g} \parallel_{2}^{2} + \lambda_{2} \parallel \boldsymbol{r} \parallel_{2}^{2} 
\end{equation}

where $\mathbf{w} \in \mathbb{R}^{m\times n}$ is the filter learned in KCF step with window size $m\times n $ and $ \mathbf{X} \in \mathbb{R}^{mn\times mn} $ is the data matrix of all possible shifts of the image patch that contains the target being tracked (like before) in which 2D convolution would be performed in the detection phase.\\ $\mathbf{r} \in \mathbb{R}^{b}$ is the rotation filter that is to be learned and $b$ is the number of bins in HOG descriptor $(\mathbf{a})$ describing globally the patch , $\mathbf{A}$ is the circulant matrix of vector $\mathbf{a}$ that reflects all possible rotations of the target, in which 1D convolution would be performed in the detection phase.$\mathbf{g}$ is just like $\mathbf{y}$ of the KCF , it is a 1D typical response of the rotation filter $\mathbf{r}$ on the $\mathbf{a}$ descriptor if there was no rotation from one frame to the other in the detection  $ \lambda_{1} , \lambda_{2} $ are the regularization variables. \\
The optimization \ref{eq:7} is separable in $\mathbf{w},\mathbf{r} $ and has the closed form solutions  :

\begin{equation}\label{eq:9}
\boldsymbol{w} = (\boldsymbol{X}^{T}\boldsymbol{X} + \lambda_{1}\boldsymbol{I} ) ^{-1} \boldsymbol{X}^{T}\boldsymbol{y}
\end{equation}

\begin{equation}\label{eq:10}
\boldsymbol{r} = (\boldsymbol{A}^{T}\boldsymbol{A} + \lambda_{2}\boldsymbol{I} ) ^{-1} \boldsymbol{A}^{T}\boldsymbol{g}
\end{equation}

Following simiar procedure as in KCF derivation we can show that $\mathbf{r}$ can be written as  :

\begin{equation}\label{eq:11}
    \boldsymbol{\hat{r}} = \frac{\boldsymbol{\hat{a}^{*}} \odot \boldsymbol{\hat{g}}}{\boldsymbol{\hat{a}^{*}} \odot \boldsymbol{\hat{a}} + \lambda_{2}} 
\end{equation}
in the dual domain we can formulate the following solution
\begin{equation}\label{eq:12}
    \boldsymbol{\hat{\alpha}_{r}} = \frac{\boldsymbol{\hat{g}}}{\boldsymbol{\hat{k}^{aa}}  + \lambda_{2}} 
\end{equation}
in which $\hat{k}^{aa} $ is just like $\hat{k}^{xx} $ in KCF. \\

\subsection{Detection phase in RKCF }
the algorithm used in detection in RKCF is similar with KCF in which we  apply the learned filter $\boldsymbol{w}$ to the current patch and we get response . comparing the position of the maximum of the response to the $\boldsymbol{y}$ maximum position . This translation of the maximum dictates the translation of the target position in the next frame . In RKCF we do the same thing as KCF then apply the $\boldsymbol{r}$ filter that we learned in\ref{eq:11} to give a response like the one in the following figure \ref{response}. The shift of the maximum of this response compared to the standard response (Gaussian centred at the middle ) will give the rotation of the object from one frame to another.

\begin{figure}[h!]
\begin{center}
   \includegraphics[width=1\linewidth]{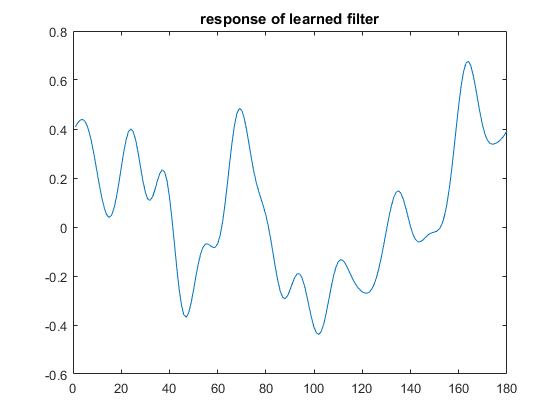}
\end{center}
   \caption{the response of the HOG of the new patch to the rotation filter ,max shift from zero represent the rotation in degrees   }
\label{response}
\end{figure}

after we have detected the rotation from one frame to the following one, this angle will be used to enhance the KCF response by counter rotate the new patch by the amount it suffered from the first frame , re apply $ \mathbf{w} $ on the "adjusted" patch and see if the max response of that was higher than the original response.If this was the case then take the new response translation as the trusted one and rotate it and apply it on the target. If not , then go with KCF suggested translation. This will insure to some extent that tracker doesn't drift away from the target based on false rotation detection. Algorithm \ref{algorithm1} summarizes the RKCF tracking scheme.\\ \\

\begin{algorithm}[h]\label{algorithm1}
 each frame, do the following:\\ 
$\bullet $ learn any CF filter , like in equation \ref{eq:5} \\
$\bullet $ apply what you learn on the new frame patch and record max response $ U $ \\
 $\bullet $ learn filter $\mathbf{r} $ based on HOG of the old patch like in \ref{eq:11} \\
$\bullet $ apply $ \mathbf{r}$ on the new patch HOG \\
$\bullet $ get the rotation $\theta$ as described in section 3 \\
$\bullet $ rotate the patch to $-\theta$ and apply the original CF on it
\uIf{$\max$(response) $> U$} {
  $\bullet $ trust the rotation and take its max translation \;
  $\bullet $ rotate the translation by $\theta$ and use it as target translation \; }
 \uElse
 { $\bullet $ use whatever $U$ gave as translation of the target\;}
\space 
 \caption{RKCF learning and detection algorithm}
\end{algorithm}

\section{Experiments}

\subsection{large scale experiment for rotation detection}
A set of 816 images were each rotated by 1000 random rotations to give a validation set of 816000 samples . To test the rotation filter effectiveness in detecting rotation, two other ways of detecting rotation based on HOG feature were tested and bench-marked. the filter have access on the upright patch and the rotated patch but no access to the actual angle at which it was rotated and its goal is to find that angle. The following results were obtained for a cosine window and a Gaussian window on the patch to cancel the boundary effect. The two other ways are correlation between the two HOGs , and observing the shift of the max of the HOG descriptor from one frame to another. The result is mean abs error in degrees for all the permutations.

\begin{table}[h!]
\centering
 \begin{tabular}{||c| c c||} 
 \hline
 boundary window & cos window & Gaussian window   \\ [0.5ex] 
 \hline\hline
  rotation filter & 15.29 & 16  \\ 
 correlation & 13.46 & 19.04  \\
 max shift & 36.96 & 43.19  \\[1ex] 
 \hline 
 \end{tabular}
\space \\
    \caption{The absolute error in degrees for different rotation detection envelops and using different rotation detection techniques based on HOG features  }
\end{table}

\subsection{The RKCF on OBT50 .}
\begin{figure}[h!]
\begin{center}
   \includegraphics[width=0.8\linewidth]{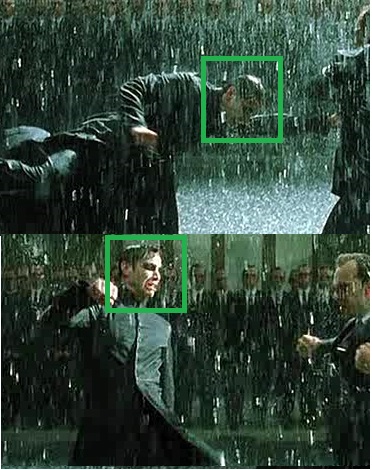}
\end{center}
   \caption{ The "Matrix" sequence in which in one frame the head is straight and five frames later its 90 degrees rotated making it difficult to track by KCF  }
\label{matrix}
\end{figure}

OBT50 is one of the most famous data-sets for visual tracking since it was released in 2013 and extended to OBT100 in 2015 \cite{Author2} .We assess our RKCF algorithm on OBT50 dataset and compare to the base line ( KCF in this case ).We observe huge enhancement of the base line for some difficult videos like the "Matrix" video on which there is a lot of rotation that a regular KCF suffers dramatically. figure \ref{matrix} show how the target suffers huge rotation from a frame to the following frame.\\

In this specific sequence the difference in precision is is 20 pts more for RKCF !. The following figure \ref{matrix_p} depicts this. 

\begin{figure}[h!]
\begin{center}
   \includegraphics[width=1\linewidth]{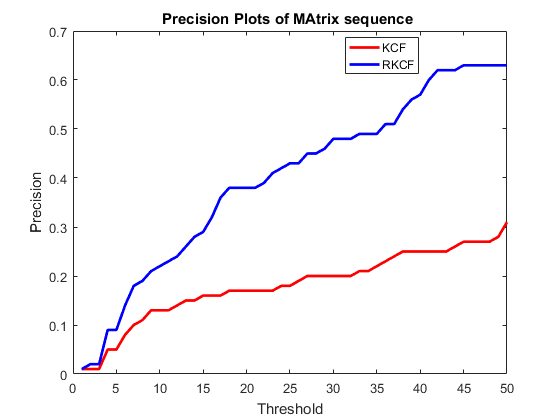}
\end{center}
   \caption{ The "Matrix" sequence precision plot comparing baseline KCF and proposed RKCF  }
\label{matrix_p}
\end{figure}

\begin{figure}[h!]
\begin{center}
   \includegraphics[width=1\linewidth]{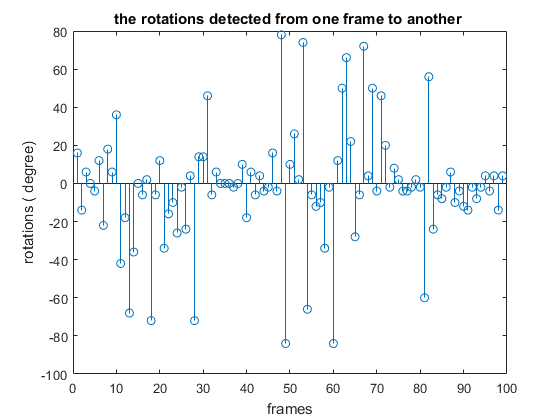}
\end{center}
   \caption{ The "Matrix" sequence detected target rotation from one frame to another   }
\label{matrix_r}
\end{figure}

We can see in this specific sequence that the rotation from one frame to another is huge as shown in figure \ref{matrix_r} . So we propose a new method to evaluate the rotational difficulty of a video directly from its KCF response sequence. We call it $\mho $ (pronounced moh) the rate of rotational change in target in which 
\begin{equation} \label{eq:13}
\mho  = std(\theta_{i}) \forall i \in n
\end{equation}
where $n$ is the number of frames in the sequence and $\theta_{i}$ is the rotation detected by RKCF in the frame $i$ . in the matrix sequence $\mho$ was 24.8 , very high rate per frame 

\begin{figure}[h!]
\begin{center}
   \includegraphics[width=1\linewidth]{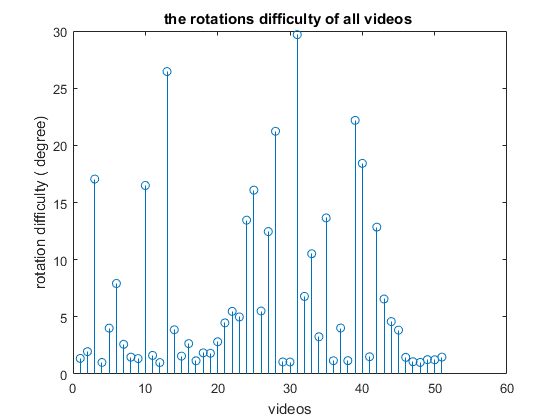}
\end{center}
   \caption{ The rotational difficulty $\mho$ on the OBT50 data set   }
\label{mho}
\end{figure}

\begin{figure}[h!]
\begin{center}
   \includegraphics[width=1\linewidth]{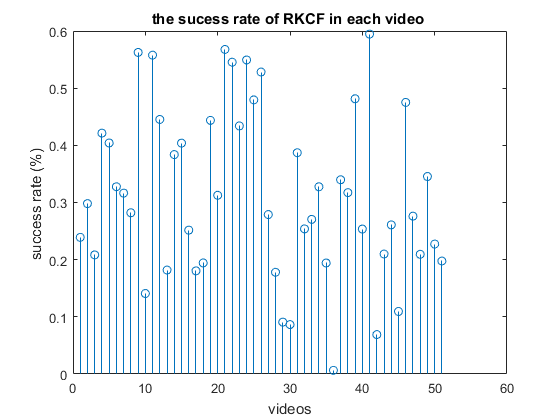}
\end{center}
   \caption{ The success rate of RKCF  $R$ on the OBT50 data set   }
\label{success}
\end{figure}

We define a success rate  $R$ for our proposed RKCF that assess its quality on a video as folows : 
\begin{equation} \label{eq:14}
R  = s/(s+f)
\end{equation}
in which $s$ is the number of frames in which proposed RKCF scheme gave higher response than baseline KCF , $f$ is the number of frames in which proposed RKCF scheme gave lower response than baseline KCF. \\
Performing the test on the whole data-set, we get the following analysis results \ref{mho}\ref{success} showing rotational difficulty $\mho$ , success rate $R$ for all videos . 

A mean success rate of all data-set of 31.57\% was obtained. The following precision plot compares RKCF and baseline KCF on the whole dataset.
\begin{figure}[h!]
\begin{center}
   \includegraphics[width=1\linewidth]{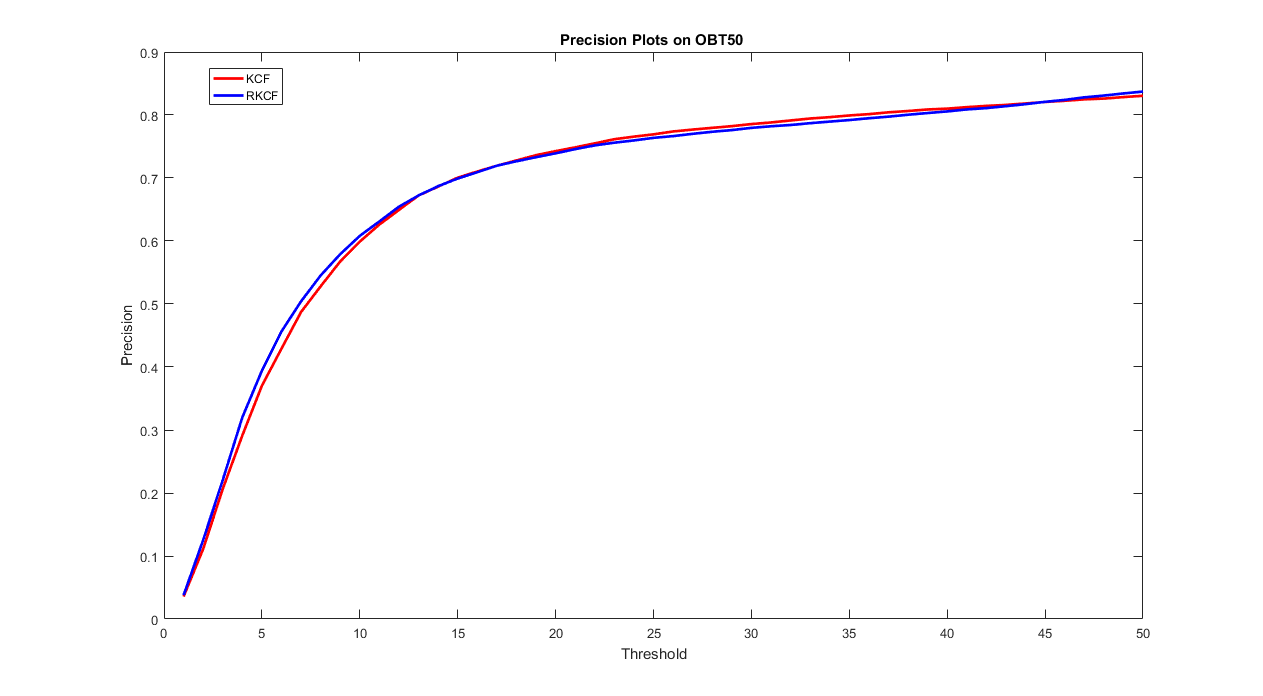}
\end{center}
   \caption{ The precision plot of RKCF  and baseline KCF on the OBT50 data set   }
\label{precision}
\end{figure}

\space

\section{Observations}
\noindent
$\bullet $  Importance of matching target size( by scale of filter) on over-all tracking precision, correct rotation with bad scale doesn't help that much  \\
$\bullet$   For constant scale targets , proposed RKCF achieve as good or better than KCF due to its rotational capability \\
$\bullet $  videos with $\mho$  aound 25 , gives max performance of RKCF over KCF.\\
$\bullet $ it might be easier and better to apply correlation between HOGs instead of learning rotational filter $\mathbf{r}$.\\

\section{Future work}
\noindent
$\bullet $  we propose generalizing this framework to include all types of features ( not only HOG) so RKCF can be used to augment Deep Features trackers like one in \cite{Author5}, the generalization can be  \\
$\bullet $we propose justification on why the typical target response of the rotational filter to be Gaussian centered at the middle .\\
$\bullet$ We propose updating the learning rule of  $\mathbf{w}$ to include the rotation information $ \theta$.\\
$\bullet $We propose a KLT framework to tackle scale and rotation in Correlation Filter fashion , by linearizing around the current frame  . \\

\newpage \clearpage
\newpage



{\small
\bibliographystyle{ieee}
\bibliography{arxive}
}

\end{document}